\documentclass{article}
\usepackage[T1]{fontenc}
\usepackage[utf8]{inputenc}
\usepackage{array}
\usepackage{refstyle}
\usepackage{booktabs}
\usepackage{url}
\usepackage{multirow}
\usepackage{amstext}
\usepackage{graphicx}
\usepackage[authoryear]{natbib}

\makeatletter

\AtBeginDocument{\providecommand\eqref[1]{\ref{eq:#1}}}
\AtBeginDocument{\providecommand\figref[1]{\ref{fig:#1}}}
\AtBeginDocument{\providecommand\tabref[1]{\ref{tab:#1}}}
\providecommand{\tabularnewline}{\\}
\RS@ifundefined{subsecref}
  {\newref{subsec}{name = \RSsectxt}}
  {}
\RS@ifundefined{thmref}
  {\def\RSthmtxt{theorem~}\newref{thm}{name = \RSthmtxt}}
  {}
\RS@ifundefined{lemref}
  {\def\RSlemtxt{lemma~}\newref{lem}{name = \RSlemtxt}}
  {}

\usepackage[final]{nips_2017}

\usepackage[utf8]{inputenc} 
\usepackage[T1]{fontenc}    
\usepackage{hyperref}       
\usepackage{url}            
\usepackage{booktabs}       
\usepackage{amsfonts}       
\usepackage{nicefrac}       
\usepackage{microtype}      

\title{To prune, or not to prune: exploring the efficacy of pruning for model compression}

\author{
  Michael H. Zhu\thanks{This research was done while the author was an intern at Google.} \\
  Department of Computer Science\\
  Stanford University\\
  Stanford, CA 94305 \\
  \texttt{mhzhu@cs.stanford.edu} \\
  \And
  Suyog Gupta \\
  Google Inc. \\
  1600 Amphitheatre Pkwy \\
  Mountain View, CA 94043 \\
  \texttt{suyoggupta@google.com} \\
}

\setcitestyle{round}
\renewcommand{\tabref}{\Tabref}
\renewcommand{\figref}{\Figref}
\usepackage{caption} 

\@ifundefined{showcaptionsetup}{}{%
 \PassOptionsToPackage{caption=false}{subfig}}
\usepackage{subfig}
\makeatother

\begin{document}
\maketitle
\begin{abstract}
Model pruning seeks to induce sparsity in a deep neural network's
various connection matrices, thereby reducing the number of nonzero-valued
parameters in the model. Recent reports \citep{DBLP:journals/corr/HanMD15,DBLP:journals/corr/NarangDSE17}
prune deep networks at the cost of only a marginal loss in accuracy
and achieve a sizable reduction in model size. This hints at the possibility
that the baseline models in these experiments are perhaps severely
over-parameterized at the outset and a viable alternative for model
compression might be to simply reduce the number of hidden units while
maintaining the model's dense connection structure, exposing a similar
trade-off in model size and accuracy. We investigate these two distinct
paths for model compression within the context of energy-efficient
inference in resource-constrained environments and propose a new gradual
pruning technique that is simple and straightforward to apply across
a variety of models/datasets with minimal tuning and can be seamlessly
incorporated within the training process. We compare the accuracy
of large, but pruned models (\emph{large-sparse}) and their smaller,
but dense (\emph{small-dense}) counterparts with identical memory
footprint. Across a broad range of neural network architectures (deep
CNNs, stacked LSTM, and seq2seq LSTM models), we find \emph{large-sparse}
models to consistently outperform \emph{small-dense} models and achieve
up to 10x reduction in number of non-zero parameters with minimal
loss in accuracy.
\end{abstract}

\section{Introduction}

Over the past few years, deep neural networks have achieved state-of-the-art
performance on several challenging tasks in the domains of computer
vision, speech recognition, and natural language processing. Driven
by increasing amounts of data and computational power, deep learning
models have become bigger and deeper to better learn from data. While
these models are typically deployed in a datacenter back-end, preserving
user privacy and reducing user-perceived query times mandate the migration
of the intelligence offered by these deep neural networks towards
edge computing devices. Deploying large, accurate deep learning models
to resource-constrained computing environments such as mobile phones,
smart cameras etc. for on-device inference poses a few key challenges.
Firstly, state-of-the-art deep learning models routinely have millions
of parameters requiring $\mathcal{O}$(MB) storage, whereas on-device
memory is limited. Furthermore, it is not uncommon for even a single
model inference to invoke  $\mathcal{O}(10^9)$ memory accesses and
arithmetic operations, all of which consume power and dissipate heat
which may drain the limited battery capacity and/or test the device’s
thermal limits. 

Confronting these challenges, a growing body of work has emerged that
intends to discover methods for compressing neural network models
while limiting any potential loss in model quality. Latency-sensitive
workloads relying on energy-efficient on-device neural network inference
are often memory bandwidth-bound, and model compression offers the
two-fold benefit of reducing the total number of energy-intensive
memory accesses as well as improving the inference time due to an
effectively higher memory bandwidth for fetching compressed model
parameters. Within the realm of model compression techniques, pruning
away (forcing to zero) the less salient connections (parameters) in
the neural network has been shown to reduce the number of nonzero
parameters in the model with little to no loss in the final model
quality. Model pruning enables trading off a small degradation in
model quality for a reduction in model size, potentially reaping improvements
in inference time and energy-efficiency. The resulting pruned model
typically has sparse connection matrices, so efficient inference using
these sparse models requires purpose-built hardware capable of loading
sparse matrices and/or performing sparse matrix-vector operations
\citep{zhang2016cambricon,DBLP:journals/corr/HanLMPPHD16,parashar2017scnn}.
Also, representing sparse matrices carries with it an additional storage
overhead increasing the model’s net memory footprint which must also
be taken into consideration.

In this work, we seek to perform a closer examination of the effectiveness
of model pruning as a means for model compression. \emph{From the
perspective of on-device neural network inference, given a bound on
the model’s memory footprint, how can we arrive at the most accurate
model?} We attempt to answer this question by comparing the quality
of the models obtained through two distinct methods: (1) training
a large model, but pruned to obtain a sparse model with a small number
of nonzero parameters (\emph{large-sparse}); and (2) training a \emph{small-dense}
model with size comparable to the \emph{large-sparse} model. Both
of these methods expose a model accuracy and size tradeoff, but differ
remarkably in terms of their implications on the design of the underlying
hardware architecture. For this comparative study, we pick models
across a diverse set of application domains: InceptionV3 \citep{szegedy2016rethinking}
and MobileNets \citep{DBLP:journals/corr/HowardZCKWWAA17} for image
recognitions tasks, stacked LSTMs for language modeling, and seq2seq
models used in Google’s Neural Machine Translation \citep{DBLP:journals/corr/WuSCLNMKCGMKSJL16}
system. In the process of this investigation, we also develop a simple
gradual pruning approach that requires minimal tuning and can be seamlessly
incorporated within the training process and demonstrate its applicability
and performance on an assortment of neural network architectures. 

\section{Related work}

Early works in the 1990s \citep{NIPS1989_250,298572} performed pruning
using a second-order Taylor approximation of the increase in the loss
function of the network when a weight is set to zero. In Optimal Brain
Damage \citep{NIPS1989_250}, the saliency for each parameter was
computed using a diagonal Hessian approximation, and the low-saliency
parameters were pruned from the network and the network was retrained.
In Optimal Brain Surgeon \citep{298572}, the saliency for each parameter
was computed using the inverse Hessian matrix, and the low-saliency
weights are pruned and all other weights in the network are updated
using the second-order information.

More recently, magnitude-based weight pruning methods have become
popular techniques for network pruning \citep{NIPS2015_5784,DBLP:journals/corr/HanMD15,DBLP:conf/conll/SeeLM16,DBLP:journals/corr/NarangDSE17}.
Magnitude-based weight pruning techniques are computationally efficient,
scaling to large networks and datasets. Our automated gradual pruning
algorithm prunes the smallest magnitude weights to achieve a preset
level of network sparsity. In contrast with the works listed above,
our paper focuses on comparing the model accuracy and size tradeoff
of \emph{large-sparse} versus \emph{small-dense} models.

A work similar to ours is the work by \citet{DBLP:journals/corr/NarangDSE17}
on pruning a RNN and GRU model for speech recognition and showing
that a sparse RNN that was pruned outperformed a dense RNN trained
normally of comparable size. While they provide one data point comparing
the performance of a sparse vs dense model, our work does an extensive
comparison of sparse vs dense models across a wide range of models
in different domains (vision and NLP). Narang et al. also introduce
a gradual pruning scheme based on pruning all the weights in a layer
less than some threshold (manually chosen) which is linear with some
slope in phase 1 and linear with some slope in phase 2 followed by
normal training. Compared to their approach, we do not have two phases
and do not have to choose two slopes, and we do not need to choose
weight thresholds for each layer (we rely on a sparsity schedule which
determines the weight thresholds). Thus, our technique is simpler,
doesn't require much hyperparameter tuning, and is shown to perform
well across different models.

Within the context of reducing model size by removing redundant connections,
several recent works \citep{DBLP:journals/corr/AnwarHS15,DBLP:journals/corr/LebedevL15,li2016pruning,DBLP:journals/corr/ChangpinyoSZ17}
propose techniques to prune and induce sparsity in a structured way,
motivated primarily by the desire to speedup computations on existing
hardware architectures optimized for dense linear algebra. Such techniques
perform coarse-grain pruning and depend critically on the structure
of the convolutional layers, and may not be directly extensible to
other neural network architectures that lack such structural properties
(LSTMs for instance). On the contrary, our method does not make any
assumptions about the structure of the network or its constituent
layers and is therefore more generally applicable.

While pruning focuses on reducing the number of non-zero parameters,
in principle, model pruning can be used in conjunction with other
techniques to further reduce model size. Quantization techniques aim
to reduce the number of bits required to represent each parameter
from 32-bit floats to 8 bits or fewer. Different quantization techniques
such as fixed-point quantization \citep{37631} or vector quantization
\citep{DBLP:journals/corr/GongLYB14} achieve different compression
ratios and accuracies but also require different software or hardware
to support inference at runtime. Pruning can be combined with quantization to achieve maximal compression \citep{DBLP:journals/corr/HanMD15}.
In addition, an emerging area of research is low precision networks
where the parameters and/or activations are quantized to 4 bits or
fewer \citep{DBLP:journals/corr/CourbariauxBD15,DBLP:journals/corr/LinCMB15,DBLP:journals/corr/HubaraCSEB16,DBLP:journals/corr/RastegariORF16,DBLP:journals/corr/ZhuHMD16}.
Besides quantization, other potentially complementary approaches to
reducing model size include low-rank matrix factorization \citep{NIPS2013_5025,NIPS2014_5544,DBLP:journals/corr/JaderbergVZ14,DBLP:journals/corr/LebedevGROL14}
and group sparsity regularization to arrive at an optimal layer size
\citep{DBLP:journals/corr/AlvarezS16}.

\section{Methods}

\begin{figure}
\begin{centering}
\includegraphics[width=0.5\linewidth]{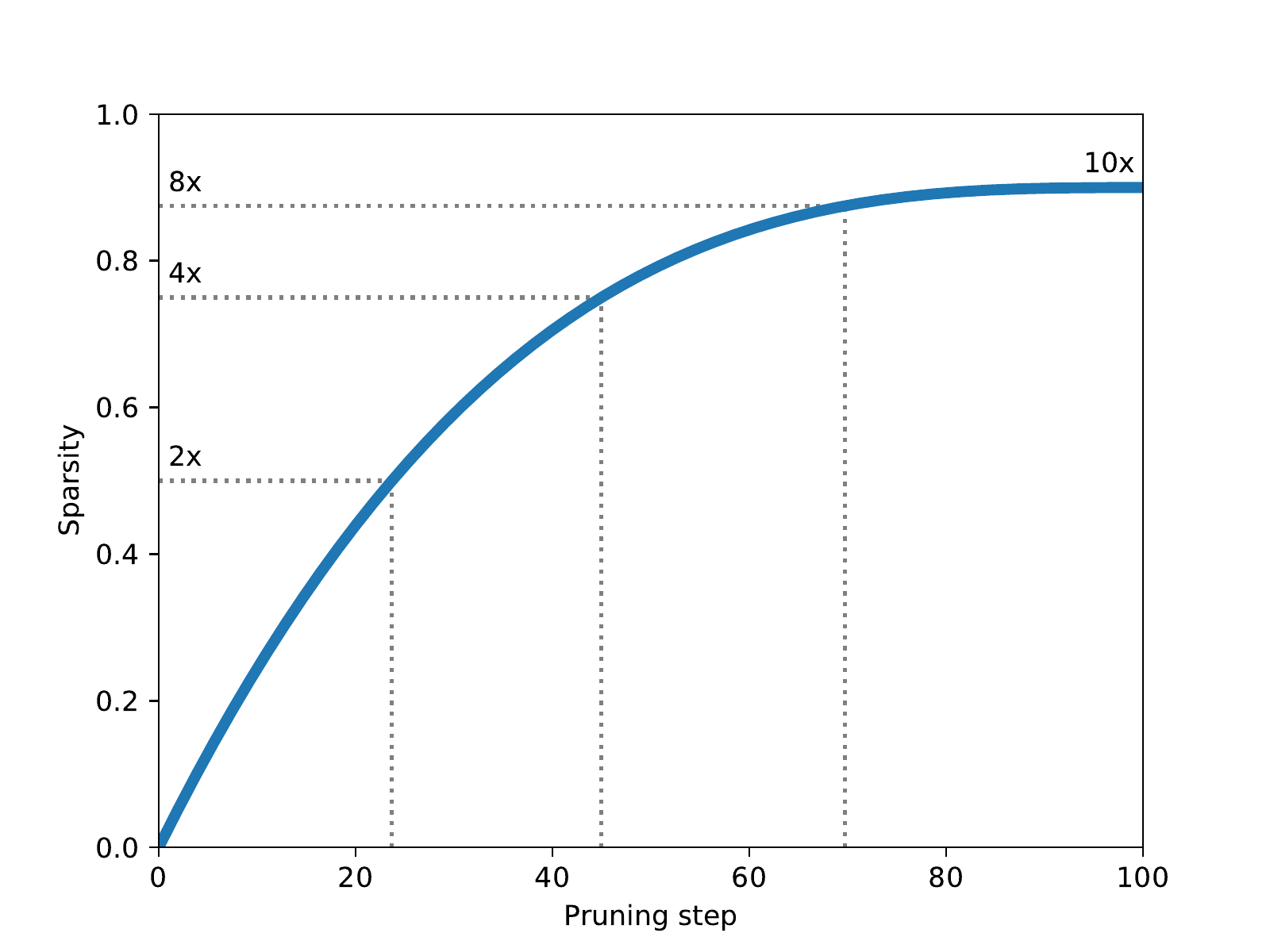}
\par\end{centering}
\caption{Sparsity function used for gradual pruning\label{fig:sparsity_function}}

\end{figure}

We extend the TensorFlow \citep{tensorflow2015-whitepaper} framework
to prune the network's connections during training. For every layer
chosen to be pruned, we add a binary mask variable which is of the
same size and shape as the layer's weight tensor and determines which
of the weights participate in the forward execution of the graph.
We inject ops into the TensorFlow training graph to sort the weights
in that layer by their absolute values and mask to zero the smallest
magnitude weights until some desired sparsity level $s$ is reached.
The back-propagated gradients flow through the binary masks, and the
weights that were masked in the forward execution do not get updated
in the back-propagation step.  We introduce a new automated gradual
pruning algorithm in which the sparsity is increased from an initial
sparsity value $s_{i}$ (usually 0) to a final sparsity value $s_{f}$
over a span of $n$ pruning steps, starting at training step $t_{0}$
and with pruning frequency $\Delta t$:

\begin{equation}
s_{t}=s_{f}+\left(s_{i}-s_{f}\right)\left(1-\frac{t-t_{0}}{n\Delta t}\right)^{3}\enskip\text{ for }\enskip t\in\left\{ t_{0},\enskip t_{0}+\Delta t,\enskip...,\enskip t_{0}+n\Delta t\right\} \label{eq:sparstiy_function}
\end{equation}

The binary weight masks are updated every $\Delta t$ steps as the
network is trained to gradually increase the sparsity of the network
while allowing the network training steps to recover from any pruning-induced
loss in accuracy. In our experience, varying the pruning frequency
$\Delta t$ between 100 and 1000 training steps had a negligible impact
on the final model quality. Once the model achieves the target sparsity
$s_{f}$, the weight masks are no longer updated. The intuition behind
this sparsity function in \eqref{sparstiy_function} is to prune the
network rapidly in the initial phase when the redundant connections
are abundant and gradually reduce the number of weights being pruned
each time as there are fewer and fewer weights remaining in the network,
as illustrated in \figref{sparsity_function}. In the experimental
results presented in this paper, pruning is initiated after the model
has been trained for a few epochs or from a pre-trained model. This
determines the value for the hyperparameter $t_{0}$. A suitable choice
for $n$ is largely dependent on the learning rate schedule. Stochastic
gradient descent (and its many variants) typically decay the learning
rate during training, and we have observed that pruning in the presence
of an exceedingly small learning rate makes it difficult for the subsequent
training steps to recover from the loss in accuracy caused by forcing
the weights to zero. At the same time, pruning with too high of a
learning rate may mean pruning weights when the weights have not yet
converged to a good solution, so it is important to choose the pruning
schedule closely with the learning rate schedule.

\begin{figure}
\centering{}\subfloat[\label{fig:inception_a}]{\centering{}\includegraphics[width=0.45\linewidth]{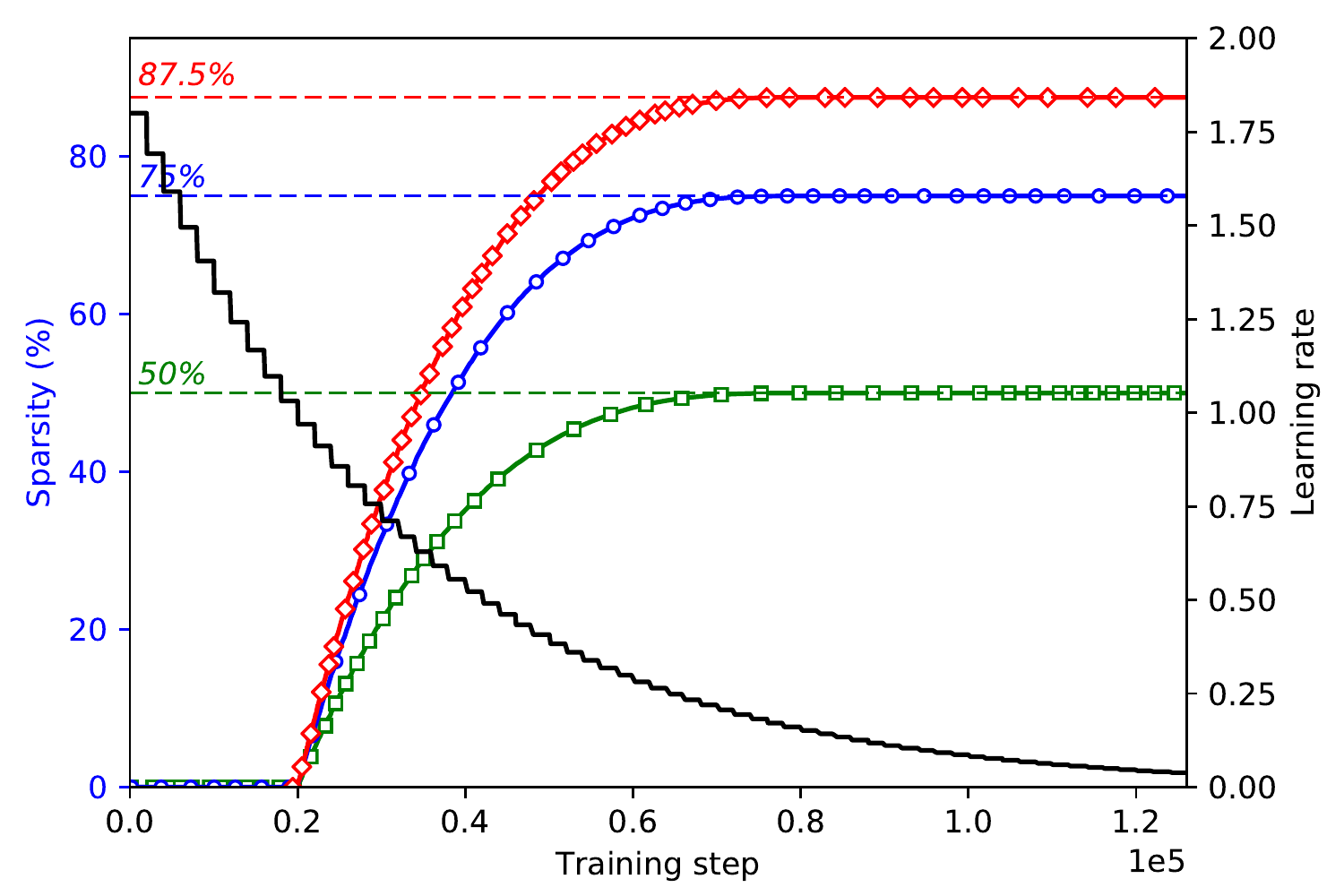}}\hspace{0.05\linewidth}\subfloat[\label{fig:inception_b}]{\begin{centering}
\includegraphics[width=0.45\linewidth]{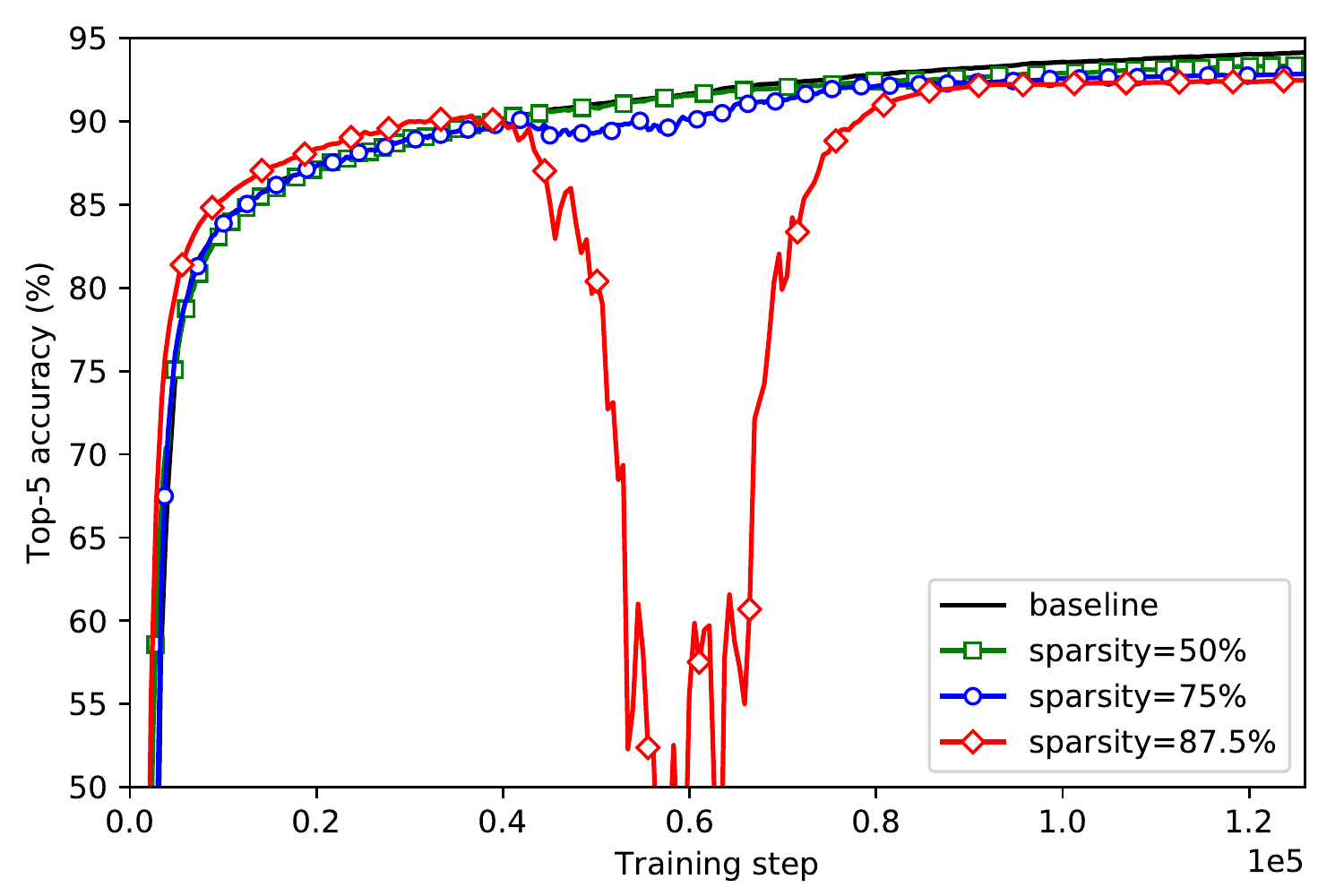}
\par\end{centering}
\centering{}}\caption{(a) The gradual sparsity function and exponentially decaying learning
rate used for training sparse-InceptionV3 models. (b) Evolution of
the model's accuracy during the training process \label{fig:inception}}
\end{figure}

\figref{inception_a} shows the learning rate and the pruning schedule
used for training sparse-InceptionV3 \citep{szegedy2016rethinking}
models. All the convolutional layers in this model are pruned using
the same sparsity function, and pruning occurs in the regime where
the learning rate is still reasonably high to allow the network to
heal from the pruning-induced damage. \figref{inception_b} offers
more insight into how this pruning scheme interacts with the training
procedure. For the 87.5\% sparse model, with the gradual increase
in sparsity, there comes a point when the model suffers a near-catastrophic
degradation, but recovers nearly just as quickly with continued training.
This behavior is more pronounced in the models trained to have higher
sparsity. \tabref{inception_table} compares the performance of sparse-InceptionV3
models pruned to varying extents. As expected, there is a gradual
degradation in the model quality as the sparsity increases. However,
a 50\% sparse model performs just as well as the baseline (0\% sparsity),
and there is only a 2\% decrease in top-5 classification accuracy
for the 87.5\% sparse model which offers an 8x reduction in number
of nonzero (NNZ) model parameters. Also note that since the weights
are initialized randomly, the sparsity in the weight tensors does
not exhibit any specific structure. Furthermore, the pruning method
described here does not depend on any specific property of the network
or the constituent layers, and can be extended directly to a wide-range
of neural network architectures. 

\begin{table}
\caption{Model size and accuracy tradeoff for sparse-InceptionV3\label{tab:inception_table}}
\centering{}%
\begin{tabular}{cccc}
\toprule 
Sparsity & NNZ params & Top-1 acc. & Top-5 acc.\tabularnewline
\midrule
0\% & 27.1M & 78.1\% & 94.3\%\tabularnewline
50\% & 13.6M & 78.0\% & 94.2\%\tabularnewline
75\% & 6.8M & 76.1\% & 93.2\%\tabularnewline
87.5\% & 3.3M & 74.6\% & 92.5\%\tabularnewline
\bottomrule
\end{tabular}
\end{table}

\section{Comparing \emph{large-sparse} and \emph{small-dense} models}

\subsection{MobileNets}

\begin{figure}
\begin{centering}
\subfloat[\label{fig:mobilenet}]{\begin{centering}
\includegraphics[width=0.5\linewidth]{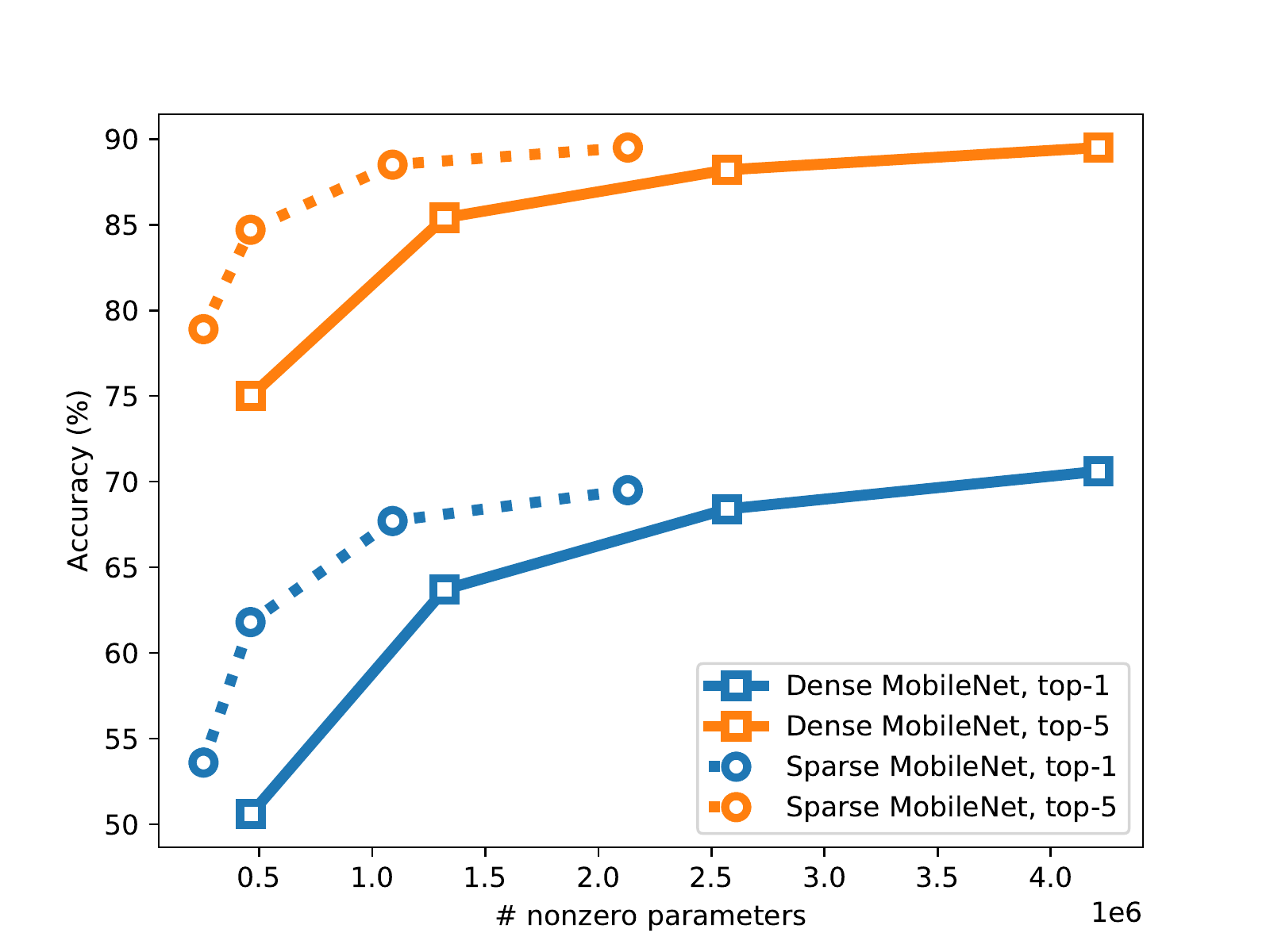}
\par\end{centering}
}\subfloat[\label{fig:ptb}]{\begin{centering}
\includegraphics[width=0.5\linewidth]{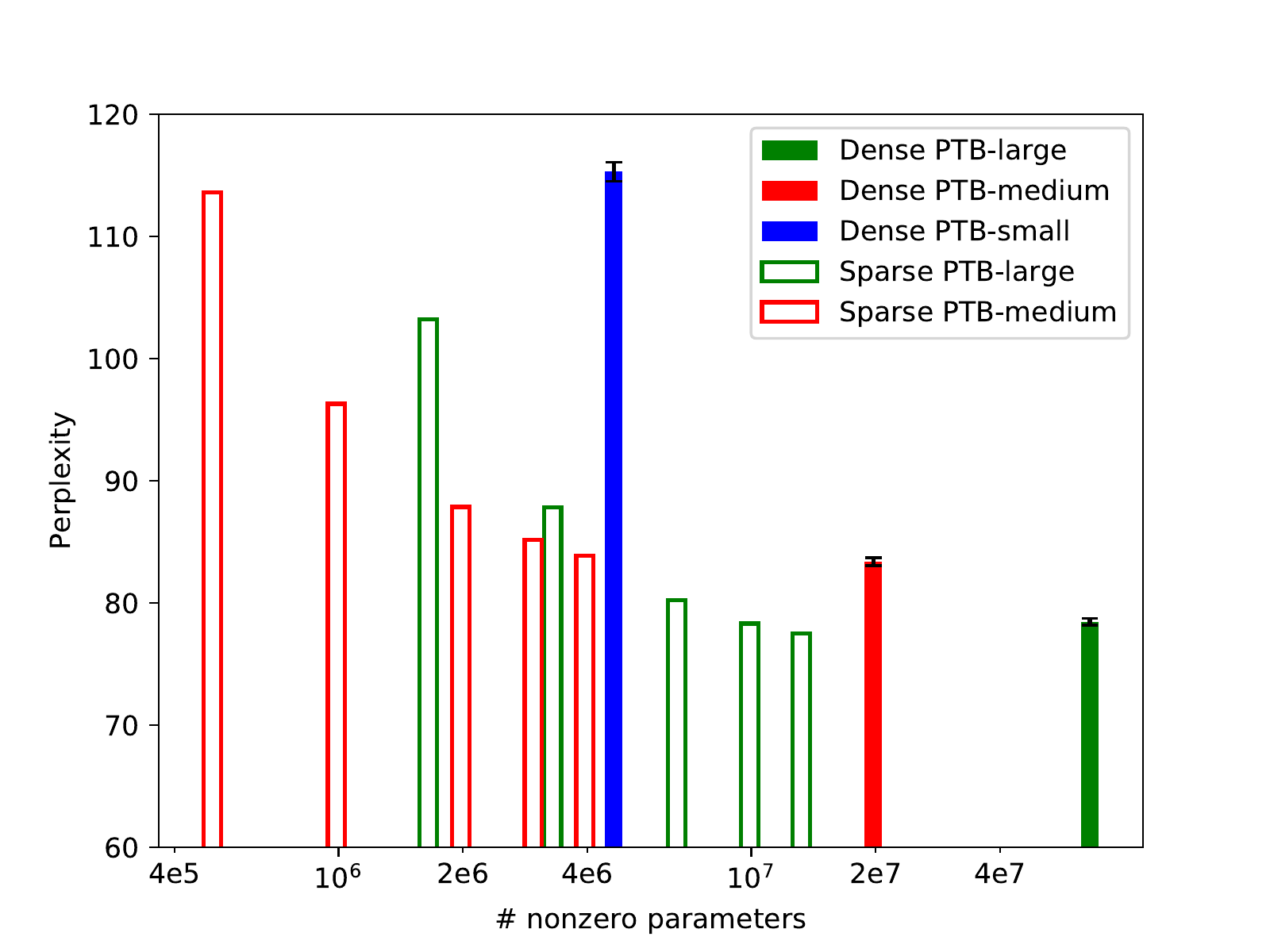}
\par\end{centering}

}
\par\end{centering}
\caption{Sparse vs dense results for (a) MobileNet and (b) Penn Treebank}
\end{figure}

\begin{table}
\caption{MobileNets sparse vs dense results\label{tab:mobilenet}}
\centering{}%
\begin{tabular}{ccccc}
\toprule 
Width & Sparsity & NNZ params & Top-1 acc. & Top-5 acc.\tabularnewline
\midrule
0.25 & 0\% & 0.46M & 50.6\% & 75.0\%\tabularnewline
\midrule
0.5 & 0\% & 1.32M & 63.7\% & 85.4\%\tabularnewline
\midrule
0.75 & 0\% & 2.57M & 68.4\% & 88.2\%\tabularnewline
\midrule
1.0 & 0\% & 4.21M & 70.6\% & 89.5\%\tabularnewline
 & 50\% & 2.13M & 69.5\% & 89.5\%\tabularnewline
 & 75\% & 1.09M & 67.7\% & 88.5\%\tabularnewline
 & 90\% & 0.46M & 61.8\% & 84.7\%\tabularnewline
 & 95\% & 0.25M & 53.6\% & 78.9\%\tabularnewline
\bottomrule
\end{tabular}
\end{table}

MobileNets are a class of efficient convolutional neural networks
designed specifically for mobile vision applications \citep{DBLP:journals/corr/HowardZCKWWAA17}.
Instead of using standard convolutions, MobileNets are based on a
form of factorized convolutions called depthwise separable convolution.
Depthwise separable convolutions consist of a depthwise convolution
followed by a 1x1 convolution called a pointwise convolution. This
factorization significantly reduces the number of parameters in the
model by filtering and combining input channels in two separate steps
instead of together as in the standard convolution. The MobileNet
architecture consists of one standard convolution layer acting on
the input image, a stack of depthwise separable convolutions, and
finally averaging pooling and fully connected layers. 99\% of the
parameters in the dense 1.0 MobileNet are in the 1x1 pointwise convolution
layers (74.6\%) and fully connected layers (24.3\%). We do not prune
the parameters in the one standard convolution layer and in the depthwise
convolution layers since there are very few parameters in these layers
(1.1\% of the total number of parameters).

The width multiplier is a parameter of the MobileNet network that
allows trading off the accuracy of the model with the number of parameters
and computational cost. The width multiplier of the baseline model
is 1.0. For a given width multiplier $\alpha\in\left(0,1\right]$,
the number of input channels and the number of output channels in
each layer is scaled by $\alpha$ relative to the baseline 1.0 model.
We compare the performance of dense MobileNets trained with width
multipliers 0.75, 0.5, and 0.25 with the performance of sparse MobileNets
pruned from dense 1.0 MobileNet in \figref{mobilenet} and \tabref{mobilenet}
on the ImageNet dataset. We see that for a given number of non-zero
parameters, sparse MobileNets are able to outperform dense MobileNets.
For example, the 75\% sparse model (which has 1.09 million parameters
and a top-1 accuracy of 67.7\%) outperforms the dense 0.5 MobileNet
(which has 1.32 million parameters and a top-1 accuracy of 63.7\%)
by 4\% in top-1 accuracy while being smaller. Similarly, the 90\%
sparse model (which has 0.46 million parameters and a top-1 accuracy
of 61.8\%) outperforms the dense 0.25 MobileNet (which has 0.46 million
parameters and a top-1 accuracy of 50.6\%) by 10.2\% in top-1 accuracy
while having the same number of non-zero parameters.

Overall, pruning is a promising approach for model compression even
for an architecture that was designed to be compact and efficient
by using depthwise separable convolutions instead of standard convolutions
as a factorization-like technique to reduce the number of parameters.
The sparsity parameter is shown to be an effective way to trade off
the accuracy of a model with its memory usage and compares favorably
with the width multiplier for MobileNet. Training a sparse MobileNet
using our gradual pruning algorithm is also easy. For pruning, we
used the same learning rate schedule as for training a dense MobileNet
but with an initial learning rate 10 times smaller than that for training
a dense MobileNet, and all other hyperparameters were kept the same.

\subsection{Penn Tree Bank (PTB) language model}

\begin{table}
\caption{PTB sparse vs dense results\label{tab:ptb}}
\centering{}%
\begin{tabular}{cccc}
\toprule 
Model & Sparsity & NNZ params & Perplexity\tabularnewline
\midrule
Small & 0\% & 4.6M & 115.30\tabularnewline
\midrule 
Medium & 0\% & 19.8M & 83.37\tabularnewline
 & 80\% & 4.0M & 83.87\tabularnewline
 & 85\% & 3.0M & 85.17\tabularnewline
 & 90\% & 2.0M & 87.86\tabularnewline
 & 95\% & 1.0M & 96.30\tabularnewline
 & 97.5\% & 0.5M & 113.6\tabularnewline
\midrule 
Large & 0\% & 66M & 78.45\tabularnewline
 & 80\% & 13.2M & 77.52\tabularnewline
 & 85\% & 9.9M & 78.31\tabularnewline
 & 90\% & 6.6M & 80.24\tabularnewline
 & 95\% & 3.3M & 87.83\tabularnewline
 & 97.5\% & 1.7M & 103.20\tabularnewline
\bottomrule
\end{tabular}
\end{table}

We train an LSTM language model on the Penn Tree Bank dataset using
the models and training procedure described in \citet{DBLP:journals/corr/ZarembaSV14}.
At each time step, the LSTM language model outputs the probability
of the next word in the sentence given the history of previous words.
The loss function is the average negative log probability of the target
words, and the perplexity is the exponential of the loss function.
The language model is composed of an embedding layer, 2 LSTM layers,
and a softmax layer. The vocabulary size is 10,000, and the LSTM hidden
layer size is 200 for the small model, 650 for the medium model, and
1,500 for the large model. In the case of the large model, there are
15M parameters in the embedding layer, 18M parameters in each of the
two LSTM layers, and 15M parameters in the softmax layer for a total
of 66M parameters. Different hyperparameters are used to train the
different-sized models. When pruning a model of a certain size, we
use the same hyperparameters that were used for training the dense
model of that size. We compare the performance of the dense models
with sparse models pruned from medium and large to 80\%, 85\%, 90\%,
95\%, and 97.5\% sparsity in \figref{ptb} and \tabref{ptb}. In this
case, we see that sparse models are able to outperform dense models
which have significantly more parameters (note the log scale for the
number of parameters). The 90\% sparse large model (which has 6.6
million parameters and a perplexity of 80.24) is able to outperform
the dense medium model (which has 19.8 million parameters and a perplexity
of 83.37), a model which has 3 times more parameters. Compared with
MobileNet, pruning PTB model likely gives better results because the
PTB model is larger with significantly more parameters. Our results
show that pruning works very well not only on the dense LSTM weights
and dense softmax layer but also the dense embedding matrix. This
suggests that during the optimization procedure the neural network
can find a good sparse embedding for the words in the vocabulary that
works well together with the sparse connectivity structure of the
LSTM weights and softmax layer.

 From \figref{ptb} and \tabref{ptb}, we also see that the 85\%
sparse medium model (which has 3 million parameters and a perplexity
of 85.17) outperforms the 95\% sparse large model (which has 3.3 million
parameters and a perplexity of 87.83). The accuracy of the 95\% sparse
large model is comparable to the accuracy of the 90\% sparse medium
model (which has 2 million parameters and a perplexity of 87.86).
Together, these results suggest that there is an optimal compression
range when pruning. In the case of PTB, pruning to 95\% sparsity for
a compression ratio of 20x significantly degrades the performance
of the sparse model compared to pruning to 90\% sparsity for a compression
ratio of 10x, as seen in \figref{ptb} from the curve of perplexity
vs. number of parameters traced by either of the sparse models. These
results suggest that in order to get the best-performing sparse model
of a certain size, we should train a dense model that is 5x-10x larger
and then prune to the desired number of parameters rather than taking
the largest and best-performing dense model and pruning this model
by 20x or more to the desired number of parameters, assuming that
the difference in performance of the two dense baseline models is
not that large. We note that it may be possible to obtain slightly
better results for pruning to 95\% sparsity or higher with more hyperparameter
tuning, and the results we obtained for pruning a model of a certain
size were from using exactly the same hyperparameter configuration
as for training the dense model of that size.

\subsection{Google Neural Machine Translation}

\begin{figure}
\begin{centering}
\subfloat[]{\begin{centering}
\includegraphics[width=0.5\linewidth]{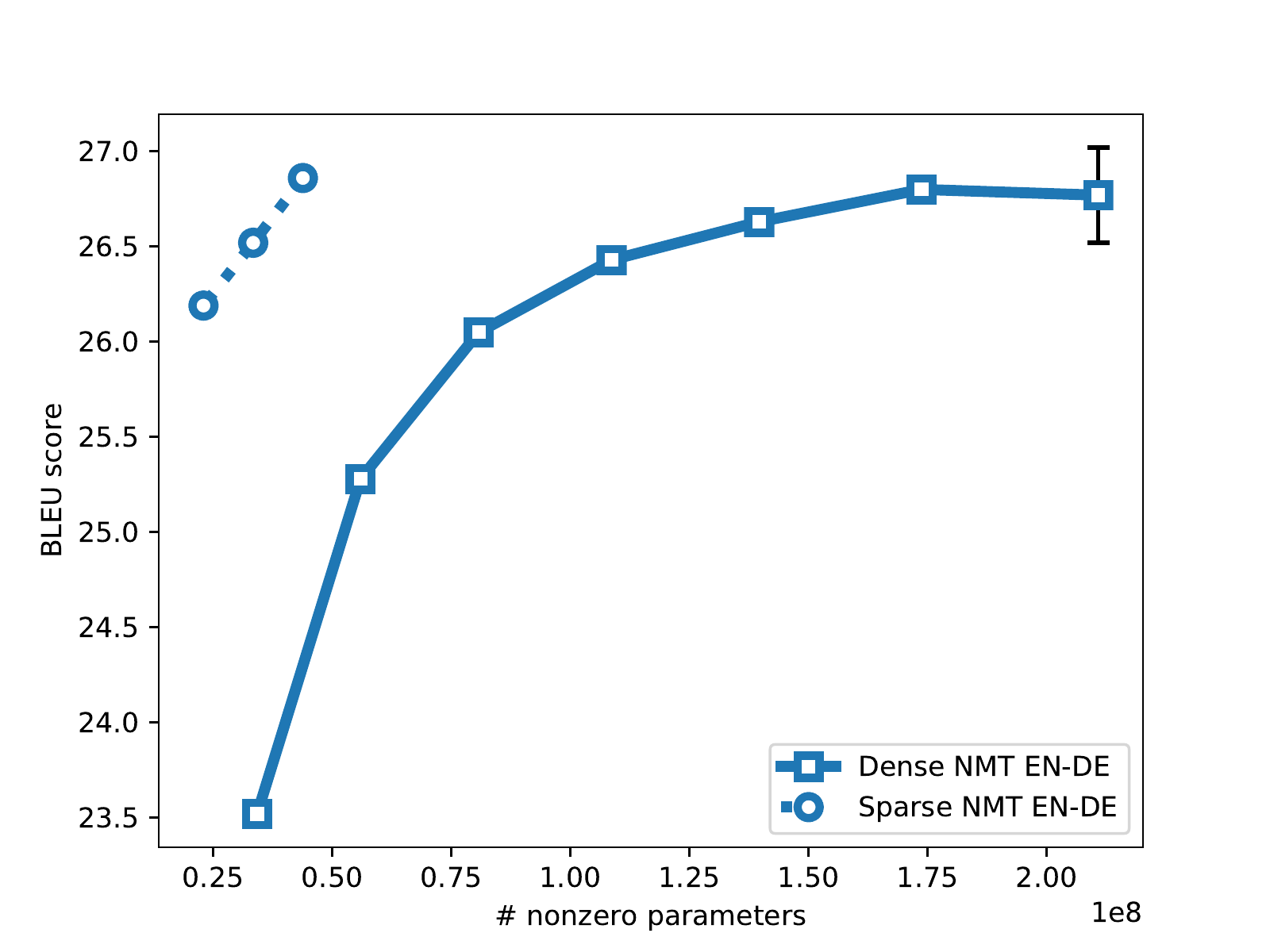}
\par\end{centering}

}\subfloat[]{\begin{centering}
\includegraphics[width=0.5\linewidth]{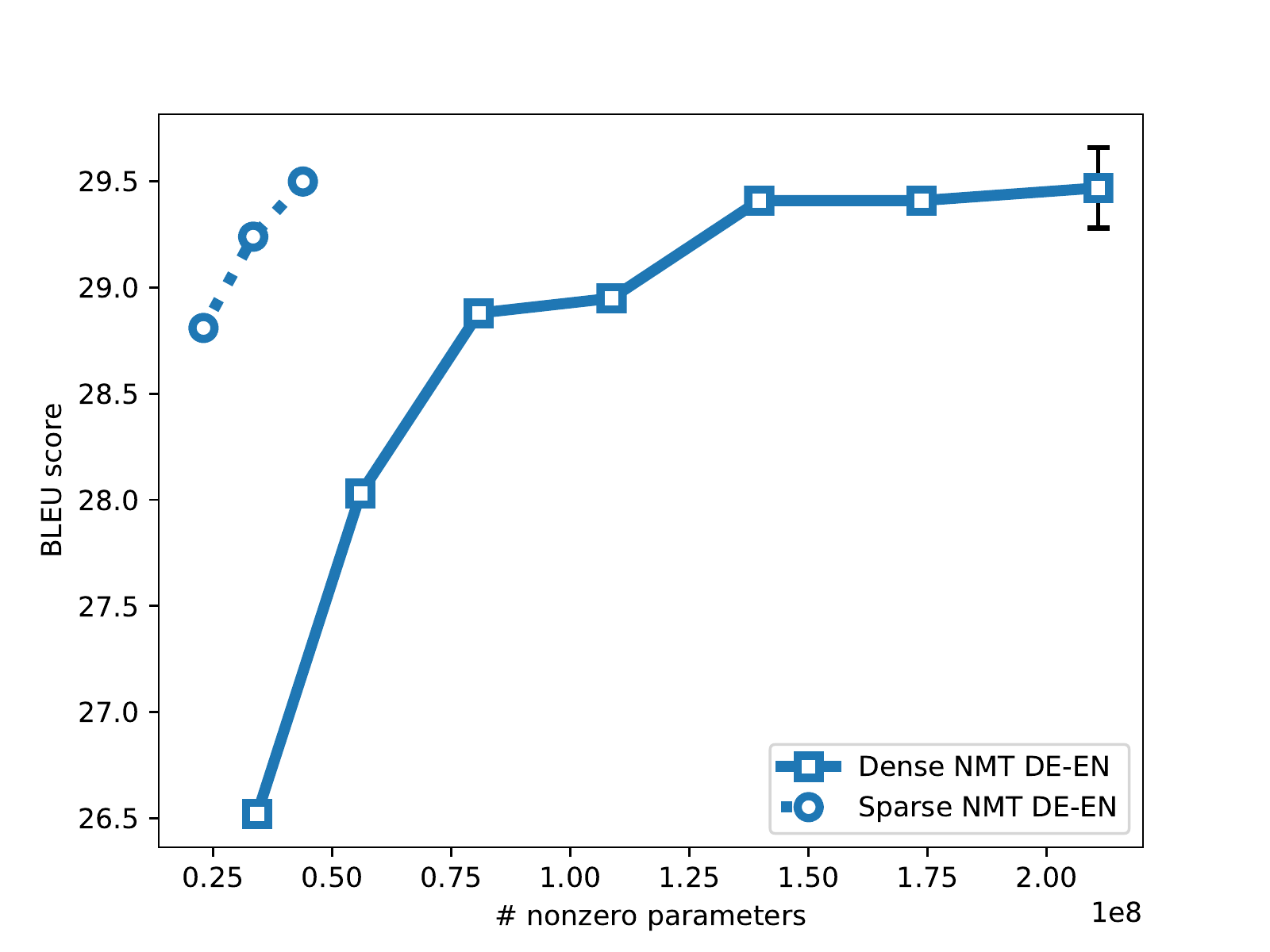}
\par\end{centering}

}
\par\end{centering}
\caption{Sparse vs dense results for (a) NMT EN-DE and (b) NMT DE-EN \label{fig:nmt}}
\end{figure}

\begin{table}
\caption{NMT sparse vs dense results \label{tab:nmt}}
\centering{}%
\begin{tabular}{ccccc}
\toprule 
\# units & Sparsity & NNZ params & EN-DE BLEU score & DE-EN BLEU score\tabularnewline
\midrule
256 & 0\% & 34M & 23.52 & 26.52\tabularnewline
\midrule
512 & 0\% & 81M & 26.05 & 28.88\tabularnewline
\midrule
768 & 0\% & 140M & 26.63 & 29.41\tabularnewline
\midrule
1024 & 0\% & 211M & 26.77 & 29.47\tabularnewline
 & 80\% & 44M & 26.86 & 29.50\tabularnewline
 & 85\% & 33M & 26.52 & 29.24\tabularnewline
 & 90\% & 23M & 26.19 & 28.81\tabularnewline
\bottomrule
\end{tabular}
\end{table}

We train a deep LSTM model for machine translation using the open-source
TensorFlow implementation available at \citet{NMT}. The implementation
is based on the Google Neural Machine Translation architecture \citep{DBLP:journals/corr/WuSCLNMKCGMKSJL16}.
The model is an encoder-decoder architecture. The encoder has an embedding
layer which maps the source vocabulary of 36,548 words into a $k$-dimensional
space, 1 bidirectional LSTM layer, and 3 standard LSTM layers. The
decoder has an embedding layer which maps the target vocabulary of
36,548 words into a $k$-dimensional space, 4 LSTM layers with attention,
and finally a softmax layer. For the dense baseline model with number
of units $k=1024$, there are 37.4M parameters in each of the encoder
embedding, decoder embedding, and softmax layers and 98.6M parameters
in all of the LSTM layers for a total of 211M parameters. Since there
are relatively few attention parameters, we do not prune the attention
parameters. We use the WMT16 German and English dataset with news-test2013
as the dev set and news-test2015 as the test set. The BLEU score is
reported as a measure of the translation quality. The learning rate
schedule used for training is 170K iterations with initial learning
rate 1.0 and 170K iterations with learning rate decay of 0.5 every
17K iterations. For pruning, the learning rate schedule we use is
70K iterations with initial learning rate 0.5 and 170K iterations
with learning decay of 0.5 every 17K iterations, and all other hyperparameters
were kept the same.

Since we noticed that the NMT training procedure had high variance,
we tested several pruning schemes applied to NMT. Our standard implementation
of gradual pruning increases the sparsity of every layer to the same
sparsity level at each pruning step. We tested a variant which we
call ``layerwise constant'' sparsity: instead of simultaneously
increasing the sparsity of all layers to some sparsity level at each
pruning step, we subdivide the pruning interval and increase the sparsity
of one layer at a time to that sparsity level. This potentially has
the effect of reducing the impact of pruning and allowing the network
to recover better with training. Finally, we compared with ``global''
pruning: we prune the smallest magnitude weights across the entire
network, regardless of which layer they are in. Global pruning produces
a different sparsity level for each layer and was shown to perform
well on NMT in the work of \citet{DBLP:conf/conll/SeeLM16}. Overall,
the layerwise constant pruning scheme performed best on average, so
we report the results with the layerwise constant pruning scheme in
\figref{nmt} and \tabref{nmt}. We note that there is high variance
in the results due to the stochasticity of the training process, as
illustrated by the error bar in \figref{nmt}. The error bar represents
the standard deviation of the BLEU score of 10 randomly initialized
and independently trained NMT models.

The results in \tabref{nmt} show that for 80\% sparsity (5x compression),
the pruned model actually achieves a slightly higher BLEU score than
the baseline model (though we note the error bar). For 85\% sparsity,
the BLEU score drops by around 0.25, and for 90\% sparsity, the BLEU
score drops by around 0.6. When we compare the performance of dense
and sparse models in \figref{nmt} and \tabref{nmt}, we again see
that sparse models outperform even larger-sized dense models. The
BLEU score of the dense model falls off quickly after 2x reduction
in model size while the BLEU score of the sparse model starts to fall
off only after 5x reduction in number of non-zero parameters. For
example, the 90\% sparse 1024-unit model is comparable to or outperforms
the dense 512-unit model (26.19 vs 26.05 for EN-DE and 28.81 vs 28.88
for DE-EN) despite having 3.5x fewer non-zero parameters (23M vs 81M).

\section{Discussion}

\begin{table}
\caption{Storage overheads associated with bit-mask and CSR(C) sparse matrix
representations for sparse-MobileNets \label{tab:overhead}}
\centering{}%
\begin{tabular}{cccc}
\toprule 
Sparsity & NNZ params & Bit-mask (MB) & CSR(C) (MB)\tabularnewline
\midrule
0\% & 4.21M & N/A & N/A\tabularnewline
50\% & 2.13M & 0.52 & 1.06\tabularnewline
75\% & 1.09M & 0.52 & 0.54\tabularnewline
90\% & 0.46M & 0.52 & 0.23\tabularnewline
95\% & 0.25M & 0.52 & 0.13\tabularnewline
\bottomrule
\end{tabular}
\end{table}

\begin{table}
\caption{Comparison of the performance of \emph{small-dense} and \emph{large-sparse}
models. Model size calculations include overhead for sparse matrix
storage and assumes 32-bit (4 bytes) per nonzero element. \label{tab:discussion-comparison}}
\centering{}%
\begin{tabular}{ccccc}
\toprule 
Model & \multicolumn{2}{c}{Small-dense} & \multicolumn{2}{c}{Large-sparse}\tabularnewline
 & Model size (MB) & Accuracy (\%) & Model size (MB) & Accuracy (\%)\tabularnewline
\cmidrule{2-5} 
\multirow{3}{*}{MobileNet} & 10.28 & 68.4 & 9.04 & \textbf{69.5}\tabularnewline
 & 5.28 & 63.7 & 4.88 & \textbf{67.7}\tabularnewline
 & \multirow{2}{*}{1.84} & \multirow{2}{*}{50.6} & 2.07 & \textbf{61.8}\tabularnewline
 &  &  & 1.13 & \textbf{53.6}\tabularnewline
\bottomrule
\end{tabular}
\end{table}

The net memory footprint of a sparse model includes the storage for
the nonzero parameters and any auxiliary data structures needed for
indexing these elements. Pruning models helps reduce the number of
nonzero-valued connections in the network; however the overhead in
sparse matrix storage inevitably diminishes the achievable compression
ratio. The bit-mask sparse matrix representation requires 1 bit per
matrix element indicating whether the element is nonzero, and a vector
containing all the nonzero matrix elements. This representation incurs
a constant overhead regardless of the model sparsity. In the compressed
sparse row (column) storage (CSR(C)) adopted in \citet{parashar2017scnn},
each nonzero parameter in the sparse matrix is associated with a count
(usually stored as a 4 or 5 bit integer) of the number of zeros preceding
it. The overhead in this case is proportional to the NNZ in the model.
\tabref{overhead} compares these two representations for sparse-MobileNets.
Naturally, the CSR(C) representation can enable higher compression
ratio for networks with high sparsity. Note, however, that the bit-mask
representation offers marginally lower overhead at smaller sparsity
levels.

In spite of this overhead, \emph{large-sparse} models appear to achieve
higher accuracy than \emph{small-dense} models with comparable memory
footprint. For instance, MobileNet with width multiplier 1 and sparsity
50\% has similar footprint as MobileNet with width multiplier 0.75,
but obtains higher accuracy. \tabref{discussion-comparison} further
highlights the trade-off between model size and accuracy for dense
and sparse models. The performance gap between \emph{large-sparse}
and \emph{small-dense} models widens for larger models such as as
the PTB language models and NMT (see \tabref{ptb} and \tabref{nmt}).
It is worth noting that the results presented in this work were obtained
by training neural networks using 32-bit floating point representation.
For neural networks trained to perform inference using reduced precision
(8-bit integer, for instance) arithmetic, the memory overhead of sparse
matrix storage represents a bigger fraction of the total memory footprint.
Quantization of the parameters to a reduced precision number representation
is also an effective method for model compression, and the interplay
between model quantization and pruning and their collective impact
on model accuracy merits a closer examination. We defer that investigation
to a future extension to this work.

\section{Conclusion}

This work sheds light on the model size and accuracy trade-off encountered
in pruned deep neural networks. We demonstrate that \emph{large-sparse}
models outperform comparably-sized \emph{small-dense} models across
a diverse set of neural network architectures. We also present a gradual
pruning technique that can be applied with ease across these different
architectures. We believe these results will encourage the adoption
of model pruning as an tool for compressing neural networks for deployment
in resource-constrained environments. At the same time, we hold the
opinion that our results will provide further impetus to the hardware
architecture community to customize the next generation of deep learning
accelerator architectures to efficiently handle sparse matrix storage
and computations.

\subsubsection*{Code}

We have open-sourced the TensorFlow pruning library used to generate
the results reported in this work.

\url{https://github.com/tensorflow/tensorflow/tree/master/tensorflow/contrib/model_pruning}

\subsubsection*{Acknowledgments}

The authors thank Huizhong Chen, Volodymyr Kysenko, David Chen, SukHwan
Lim, Raziel Alvarez, and Thang Luong for helpful discussions.

\bibliographystyle{plainnat}
\bibliography{references}

\end{document}